\documentclass[10pt,twocolumn,letterpaper]{article}

\usepackage[pagenumbers]{wacv} 

\usepackage{array}
\usepackage{algorithm}
\usepackage{algpseudocode}
\usepackage{multirow}
\usepackage{csquotes}

\definecolor{wacvblue}{rgb}{0.21,0.49,0.74}
\usepackage[pagebackref,breaklinks,colorlinks,allcolors=wacvblue]{hyperref}

\def\wacvPaperID{*****} 
\def\confName{WACV}
\def\confYear{2026}

\title{SOLAR: Switchable Output Layer for Accuracy and Robustness in Once-for-All Training}

\author{%
  Shaharyar Ahmed Khan Tareen \\
  University of Houston\\
  Houston, TX, USA \\
  \tt{stareen@cougarnet.uh.edu}   \and
  Lei Fan \\
  University of Houston\\
  Houston, TX, USA \\
  \tt{lfan8@central.uh.edu}   \and
  Xiaojing Yuan \\
  University of Houston\\
  Houston, TX, USA \\
  \tt{xyuan@uh.edu}   \and
  Qin Lin \\
  University of Houston\\
  Houston, TX, USA \\
  \tt{qlin21@central.uh.edu}   \and
  Bin Hu \\
  University of Houston\\
  Houston, TX, USA \\
  \tt{bhu11@central.uh.edu}
}

\begin{document}
\maketitle
\begin{abstract}
Once-for-All (OFA) training enables a single super-net to generate multiple sub-nets tailored to diverse deployment scenarios, supporting flexible trade-offs among accuracy, robustness, and model-size without retraining. However, as the number of supported sub-nets increases, excessive parameter sharing in the backbone limits representational capacity, leading to degraded calibration and reduced overall performance. To address this, we propose \textbf{SOLAR} (\textbf{\underline{S}}witchable \textbf{\underline{O}}utput \textbf{\underline{L}}ayer for \textbf{\underline{A}}ccuracy and \textbf{\underline{R}}obustness in Once-for-All Training), a simple yet effective technique that assigns each sub-net a separate classification head. By decoupling the logit learning process across sub-nets, the Switchable Output Layer (SOL) reduces representational interference and improves optimization, without altering the shared backbone. We evaluate SOLAR on five datasets (SVHN, CIFAR-10, STL-10, CIFAR-100, and TinyImageNet) using four super-net backbones (ResNet-34, WideResNet-16-8, WideResNet-40-2, and MobileNetV2) for two OFA training frameworks (OATS and SNNs). Experiments show that SOLAR outperforms the baseline methods: compared to OATS, it improves accuracy of sub-nets up to \textbf{1.26\%}, \textbf{4.71\%}, \textbf{1.67\%}, and \textbf{1.76\%}, and robustness up to \textbf{9.01\%}, \textbf{7.71\%}, \textbf{2.72\%}, and \textbf{1.26\%} on SVHN, CIFAR-10, STL-10, and CIFAR-100, respectively. Compared to SNNs, it improves TinyImageNet accuracy by up to \textbf{2.93\%}, \textbf{2.34\%}, and \textbf{1.35\%} using ResNet-34, WideResNet-16-8, and MobileNetV2 backbones (with 8 sub-nets), respectively.
\end{abstract}
    
\section{Introduction}
\label{sec:intro}
Deploying deep neural networks across a wide range of devices—from high-performance servers to
resource-constrained edge platforms—requires customized models that balance accuracy, robustness \cite{zhang_towards_2021, goodfellow_explaining_2015,nguyen_deep_2015,carlini_towards_2017}, and model-size (or efficiency). Once-for-All (OFA) training \cite{cai_once-for-all:_2020, chen2021onlytrainonce, yu2018slimmable, yu2019universally, kundu_float:_2023, kundu2023sparse} addresses this by yielding a single versatile super-network containing many sub-networks that are tailored to different deployment constraints. The sub-nets are selected post-training to meet trade-offs among accuracy, adversarial robustness \cite{zhang_theoretically_2019, madry_towards_2019, chen2020adversarial}, model-size, or computational cost, without retraining from scratch \cite{cai_once-for-all:_2020, yu2018slimmable, yu2019universally, kundu_float:_2023, wang_once-for-all_2020}. While OFA training \cite{yu2018slimmable, yu2019universally,wang_once-for-all_2020} offers flexibility and efficiency, scaling to a large number of sub-nets introduces a fundamental challenge: \textit{excessive parameter sharing}. When all sub-nets share a single output layer, representational interference occurs, preventing each sub-net from optimizing independently. This coupling of parameters degrades accuracy, calibration, and robustness, particularly for sub-nets with differing capacities or architectures.

\begin{figure*}[!t]
    \centering    \includegraphics[width=0.8\linewidth]{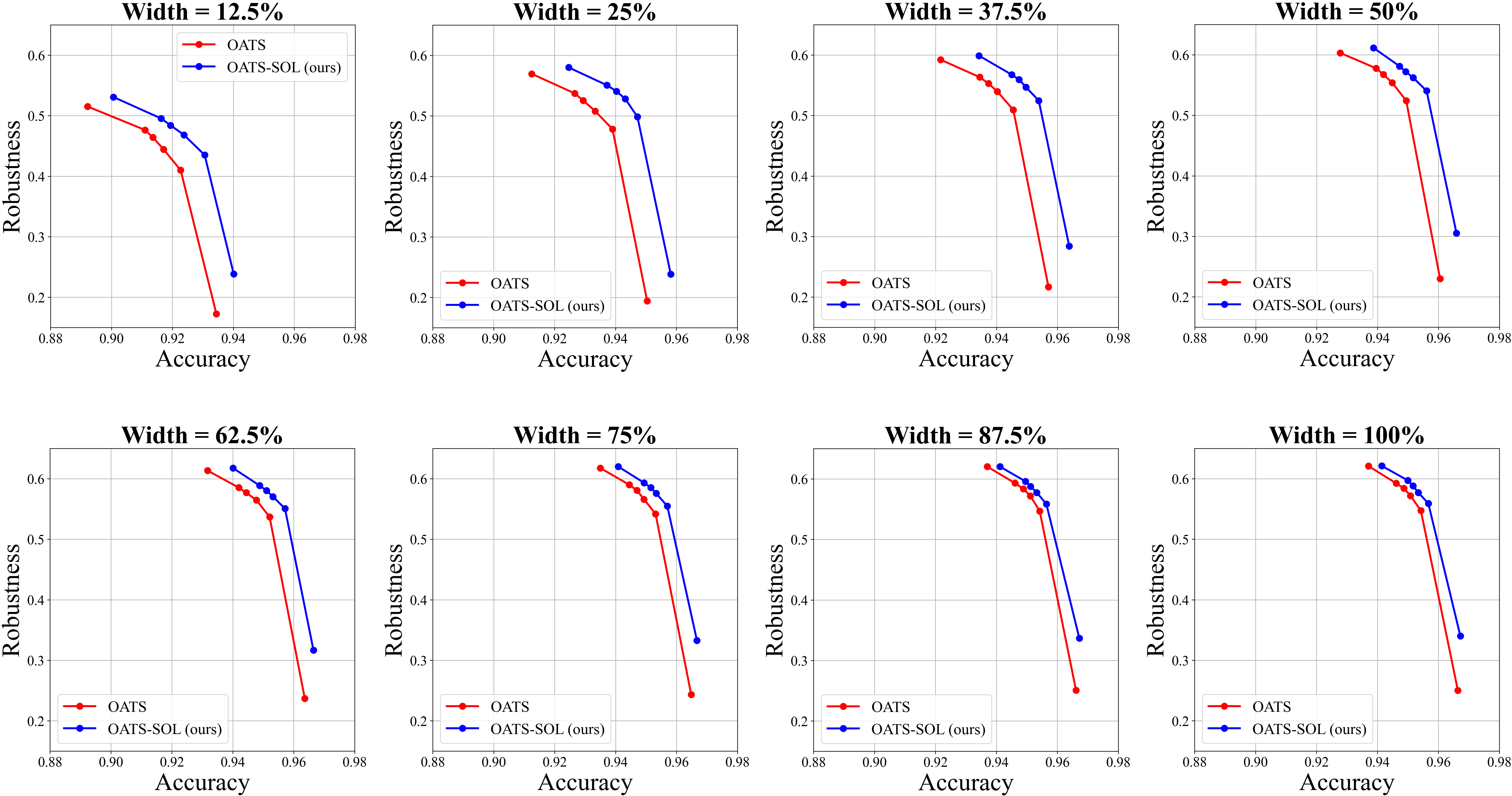}
    \caption{\centering Comparison of OATS \cite{wang_once-for-all_2020} and OATS-SOL on SVHN dataset using WideResNet-16-8 backbone packed with 8 sub-nets. OATS-SOL provides superior performance than OATS for all the sub-nets in terms of accuracy and PGD-7 robustness.}
    \label{fig:fig_1}
\end{figure*}

In this paper, we identify the shared output layer as a bottleneck in OFA frameworks and propose
\textbf{SOLAR} (\textit{Switchable Output Layer for Accuracy and Robustness in Once-for-All Training}), a simple yet effective approach that introduces separate classification heads for the sub-nets. SOLAR decouples the logit learning process in the common output layer, mitigating logit interference during training and improving the sub-net specific optimization while maintaining the training efficiency.

\textbf{Key Contributions:} Our main contributions are summarized below:

\begin{itemize}
    \item We identify the shared output layer as a bottleneck in OFA training, that leads to representational interference and performance degradation across the sub-nets. To address this, we propose \textbf{SOLAR} (\textit{Switchable Output Layer for Accuracy and Robustness}), a simple and effective method that assigns each sub-net a separate classification head while preserving the shared backbone.
    
    \item We incorporate SOLAR into two OFA frameworks: Slimmable Neural Networks (SNNs) \cite{yu2018slimmable}, which vary network width dynamically during standard training, and Once-for-All Adversarial Training and Slimming (OATS) \cite{wang_once-for-all_2020}, which combines adversarial training with the dynamic width shrinking and uses conditional loss function.
    
    \item We perform extensive experiments across five benchmark datasets (SVHN \cite{netzer2011reading}, CIFAR-10 \cite{krizhevsky2009learning}, STL-10 \cite{coates2011analysis}), CIFAR-100 \cite{krizhevsky2009learning}, and TinyImageNet \cite{le2015tiny}, using four different super-net backbones (WideResNet-16-8 \cite{zagoruyko2016wideRN}, ResNet-34 \cite{he2016deepRN}, WideResNet-40-2 \cite{zagoruyko2016wideRN}, MobileNetV2 \cite{sandler2018mobilenetv2}), demonstrating that SOLAR generalizes well and improves both standard accuracy and adversarial robustness across the sub-nets and frameworks.

    \item Our smallest sub-net, trained using OATS-SOL on the SVHN dataset using WideResNet-16-8 backbone, achieves the best accuracy of \textbf{94.01\%} and robustness of \textbf{53.08\%}, surpassing the standard OATS \cite{wang_once-for-all_2020} baseline by \textbf{0.57\%} and \textbf{1.57\%}, respectively, with a compact model size of \textbf{387 KB}.
\end{itemize}
\section{Related Work}
\label{sec:related}

\begin{figure*}[!t]
    \centering    \includegraphics[width=0.7\linewidth]{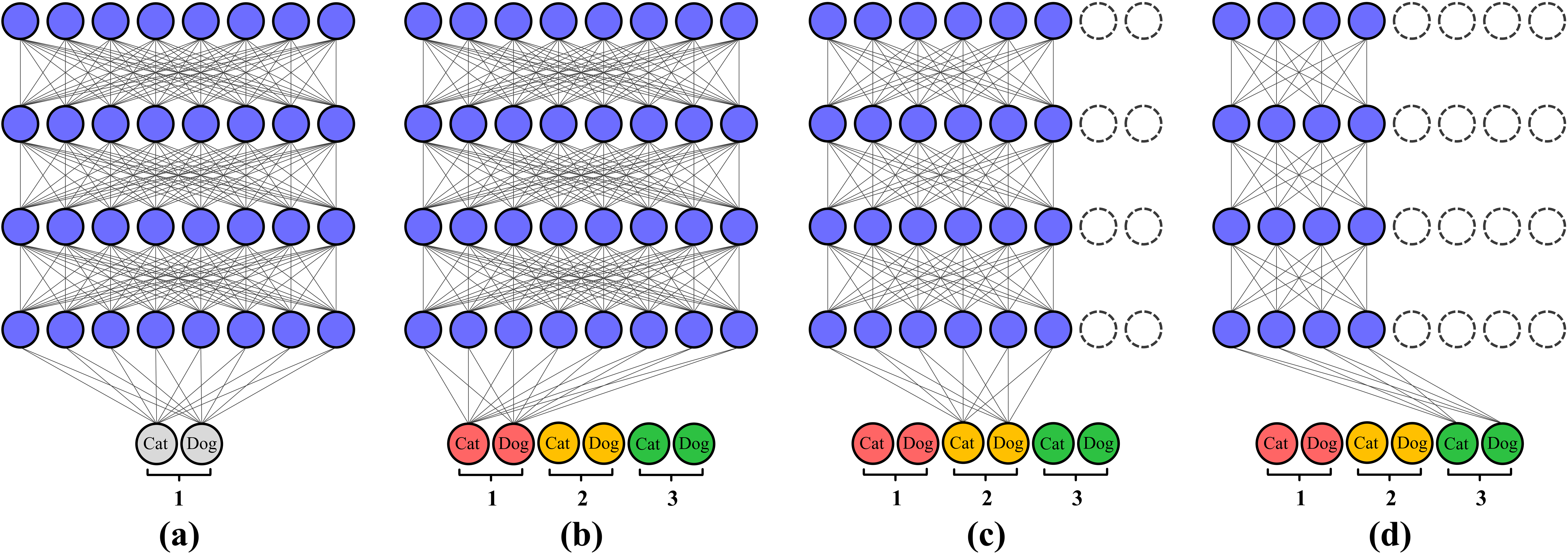}
    \caption{\centering Illustration of a vanilla SNN \cite{yu2018slimmable} vs. SNN with Switchable Output Layer (SNN-SOL) in a super-net backbone with three sub-nets of different widths: (a) vanilla SNN with a shared output layer; (b–d) SNN-SOL with 100\%, 75\%, and 50\% widths, respectively. Width refers to the number of channels per layer. SOL adds separate classification head for each sub-net, enabling decoupled logit learning, with the number of heads equal to the number of sub-nets in the backbone.}
    \label{fig:fig_2}
\end{figure*}

\textbf{Once-for-All (OFA) Training:\quad} OFA framework \cite{cai_once-for-all:_2020} trains a single over-parameterized super-net from which many sub-nets can be derived by sampling architectures with different depths, widths, kernel sizes, or input resolutions. These sub-nets inherit weights from the super-net, enabling efficient deployment without retraining from scratch. A progressive shrinking strategy \cite{cai_once-for-all:_2020, chen2021onlytrainonce, pan2023progressive} is used to jointly optimize all sub-nets. Although OFA enables massive scalability, supporting over $10^{19}$ sub-nets, training a super-net that performs well across all sub-nets is hard, because smaller sub-nets suffer from degraded performance due to conflicting gradients and shared parameters. When many sub-nets share layers, gradients from different sub-nets can interfere, making it harder to optimize the shared weights for all sub-net configurations \cite{yu2018slimmable, wang_once-for-all_2020}.

\textbf{Slimmable Neural Networks (SNNs):\quad} SNNs \cite{yu2018slimmable} follow the OFA principle by training a single super-net operating only at four widths (0.25×, 0.5×, 0.75×, 1.0×). SNNs provide a twofold trade-off between accuracy and model size (or efficiency). They address key OFA challenges—particularly performance degradation caused by conflicting feature statistics when all sub-nets share a single Batch Normalization (BN) layer \cite{ioffe2015batch}—by introducing Switchable Batch Normalization (SBN) \cite{yu2018slimmable}, which assigns separate BN layers to each sub-net. This design reduces training instability and gradient interference, improving fairness across widths \cite{galloway_batch_2019}. However, SNNs still rely on a shared output layer for all sub-nets, which becomes a bottleneck as the sub-net diversity grows. This limits the capacity to fully adapt output representations to varying sub-net complexities. Our proposed Switchable Output Layer (SOL) solves this problem by providing sub-net-specific classification heads, effectively overcoming the common output layer bottleneck and further enhancing sub-net performance without sacrificing training cost.

\textbf{Once-for-All Adversarial Training and Slimming (OATS):\quad} OATS \cite{wang_once-for-all_2020} extends the Once-for-All Adversarial Training (OAT) framework \cite{wang_once-for-all_2020} by integrating model compactness across widths like the SNNs \cite{yu2018slimmable}. It trains a single super-net supporting three widths (0.5×, 0.75×, 1.0×) via channel-wise slimming, enabling deployment across devices with varying resources. During training, OATS conditions on both adversarial loss weight $\lambda$ and width fraction, allowing the super-net to enable a balance between accuracy, robustness, and efficiency without requiring retraining from scratch. To handle distribution mismatches between clean and adversarial samples \cite{galloway_batch_2019}, OATS introduces Switchable Dual Batch Normalization (SDBN) \cite{wang_once-for-all_2020} with separate BN layers for each data type, and sub-net width, ensuring stable, high-performance training. Conditional learning techniques such as FiLM layers \cite{perez2018film, wang_once-for-all_2020} or scaled noise injection \cite{kundu_float:_2023, kundu2023sparse} enable adaptive behavior based on input conditions. OATS \cite{wang_once-for-all_2020} employs FiLM layers for this purpose. Our proposed Switchable Output Layer (SOL) improves the performance of both SNNs and OATS frameworks across diverse datasets and architectures, offering superior accuracy, robustness, and efficiency trade-offs.
\section{Preliminaries}
\label{preliminaries}

Consider a multi-class classification setting with \( N \) training samples and \( C \) classes. For each sample \( i \), let \( y_i \in \{1, \dots, C\} \) denote the ground-truth label, and let \( \mathbf{p}_i = (p_{i1}, \dots, p_{iC}) \) denote the predicted class probabilities, computed via the softmax function from the model logits \( z_{ic} \):
\begin{equation}
p_{ic} = \frac{\exp(z_{ic})}{\sum_{j=1}^{C} \exp(z_{ij})}
\label{eq:eq_1}
\end{equation}
Given a dataset \( \mathcal{D} = \{(\mathbf{x}_i, y_i)\}_{i=1}^N \), where \( \mathbf{x}_i \in \mathbb{R}^d \) and \( y_i \in \{1, \dots, C\} \), a deep neural network (DNN) \( f : \mathbb{R}^d \to \mathbb{R}^C \) with parameters \( \theta \) maps inputs to logits. The model is typically trained using empirical risk minimization (ERM) with the cross-entropy loss:
\begin{equation}
    \mathcal{L}_{\mathrm{CE}} =\mathcal{L}(f(\mathbf{x}_i; \theta), y_i) = -\log p_{i y_i}
\label{eq:eq_2}
\end{equation}

\textbf{Adversarial Training (AT): \quad} AT has been widely adopted to improve model robustness by explicitly optimizing for performance under worst-case input perturbations \cite{goodfellow_explaining_2015, madry_towards_2019, chen2020adversarial,zhao2022adversarial, huang2020bridging, shafahi2019adversarial, bai2021recent, yinusa2024evaluating}. A common approach is to use a hybrid loss \( \mathcal{L}_{\mathrm{Hybrid}} \) that combines standard classification loss \( \mathcal{L}_{\mathrm{CE}} \) and adversarial loss \( \mathcal{L}_{\mathrm{ADV}} \) \cite{zhang_theoretically_2019, wang2019improving, tsipras2018robustness, cai2018curriculum}:
\begin{equation}
    \min_{\theta} \mathbb{E}_{(\mathbf{x}, y) \sim \mathcal{D}} \underbrace{\left[ (1 - \lambda) \mathcal{L}_{\mathrm{CE}} + \lambda \mathcal{L}_{\mathrm{ADV}} \right]}_{\mathcal{L}_{\mathrm{Hybrid}}}
\label{eq:eq_3}
\end{equation}
where the adversarial loss is defined as:
\begin{equation}
    \mathcal{L}_{\mathrm{ADV}} = \max_{\delta \in \mathcal{B}_\epsilon(\mathbf{x})} \mathcal{L}(f(\mathbf{x} + \delta; \theta), y)
\label{eq:eq_4}
\end{equation}
and \( \mathcal{B}_\epsilon(\mathbf{x}) = \{ \delta \in \mathbb{R}^d \mid \|\delta\|_\infty \leq \epsilon \} \) is the \( L_\infty \)-ball around \( \mathbf{x} \), constraining the magnitude of adversarial perturbations. A widely used method for solving the inner maximization problem in adversarial training is Projected Gradient Descent~(PGD) \cite{madry_towards_2019, huang2020bridging, cai2018curriculum}. Given an input \( \mathbf{x} \) and label \( y \), PGD generates an adversarial example \( \mathbf{x}' \) as:
\begin{equation}
    \mathbf{x}' = \Pi_{\mathcal{B}_\epsilon(\mathbf{x})} \left( \mathbf{x} + \gamma \cdot \text{sign} \left( \nabla_{\mathbf{x}} \mathcal{L}(f(\mathbf{x}; \theta), y) \right) \right)
\label{eq:eq_5}
\end{equation}
where \( \gamma \) is the step size and \( \Pi_{\mathcal{B}_\epsilon(\mathbf{x})}(\cdot) \) denotes projection onto the \( L_\infty \)-ball around \( \mathbf{x} \). This procedure can be iterated multiple times to find stronger adversarial examples.

\textbf{Switchable Dual Batch Normalization (SDBN):\quad} Dual Batch Normalization~(DBN) \cite{ioffe2015batch,xie_intriguing_2019,ding_sensitivity_2019} addresses the distribution mismatch between clean and adversarial inputs during adversarial training. Instead of a single BN layer, DBN maintains two sets of BN statistics: one for clean data (\( \mathrm{BN}_{cln} \)) and one for adversarial data (\( \mathrm{BN}_{adv} \)) \cite{galloway_batch_2019, ding_sensitivity_2019}. Such normalization strategy prevents feature distortion and improves performance and robustness \cite{galloway_batch_2019,bjorck2018understanding,wang2022removing}. SDBN extends this idea to super-nets containing multiple sub-nets and maintains DBN layer for each switch (or sub-net) in the backbone \cite{wang_once-for-all_2020}. Each sub-net in the width adaptive backbones \cite{yu2018slimmable, wang_once-for-all_2020}, exhibits distinct feature statistics \cite{yu2018slimmable, wang_once-for-all_2020}. SNNs use SBN to preserve performance across all widths, that maintains separate BN parameters \((\gamma_{\alpha_k}, \beta_{\alpha_k}, \mu_{\alpha_k}, \sigma^2_{\alpha_k})\), leading to the following parameter set.
\begin{equation}
    \Theta = \left\{ W, \{ \gamma_{\alpha_k}, \beta_{\alpha_k}, \mu_{\alpha_k}, \sigma^2_{\alpha_k} \}_{k=1}^K \right\}
\label{eq:eq_6}
\end{equation}
Where \( W \) represents the shared weights of Convolutional and Linear layers, \( \gamma_{\alpha_k}, \beta_{\alpha_k} \) are the scale and shift parameters, \( \mu_{\alpha_k}, \sigma^2_{\alpha_k} \) are the running mean and variance for each sub-net with width fraction $\alpha_k$. These parameters are not shared among the sub-nets. Beyond normalization, OATS \cite{wang_once-for-all_2020} uses model-conditional learning to enable on-the-fly accuracy–robustness trade-offs by conditioning the model on structural or stochastic inputs~(e.g. via FiLM layers on a hyperparameter like \( \lambda \)~) \cite{wang_once-for-all_2020,huang2017multi, kaya2019shallow, wang2018skipnet}. Despite these solutions, a bottleneck remains in the end: the \textbf{shared output layer} where all sub-nets share a single classification head, causing \emph{logit interference} that limits learning capacity and calibration. To address this, we introduce the \textbf{Switchable Output Layer (SOL)}, which assigns a separate classification head to each sub-net in the backbone. Like SBN separates out the feature normalization for different sub-nets \cite{yu2018slimmable}, the SOL isolates their logit learning processes, enabling overall better optimization for each sub-net.
\section{Methodology}
\label{methodology}

\subsection{Bottleneck in OFA Training: Standard Output Layer (Shared Parameters)}
OFA training aims to train a single over-parametrized super-net from which multiple sub-nets can be efficiently derived for diverse deployment requirements without retraining. Let $\mathcal{N}(x; \Theta)$ denote the super-net with input \( x\) and the parameter set \( \Theta \), as defined in \eqref{eq:eq_6}. To enable flexible sub-net instantiation, a predefined set of width multipliers $\mathcal{W} = \{ \alpha_1, \alpha_2, \dots, \alpha_k \}
\quad \text{where} \quad 0 < \alpha_1 < \alpha_2 < \dots < \alpha_k \leq 1$ is used to scale the number of active channels in each layer of $\mathcal{N}$ \cite{yu2018slimmable, wang_once-for-all_2020}. The \(k^{th}\) sub-net, corresponding to width multiplier \( \alpha_k \), is denoted as $\mathcal{S}_{\alpha_k}(x) = \mathcal{N}(x; \Theta_{\alpha_k})$ where \( \Theta_{\alpha_k} \subseteq \Theta \) represents the subset of parameters utilized by the sub-net. The full-width super-net corresponds to \( \alpha_K = 1\), using the complete parameter set $\Theta$. The goal of OFA training is to learn optimal parameters \( \Theta \) that minimize a predefined objective as in \eqref{eq:eq_7}.

\begin{figure}[t]
  \centering  \includegraphics[width=0.7\linewidth]{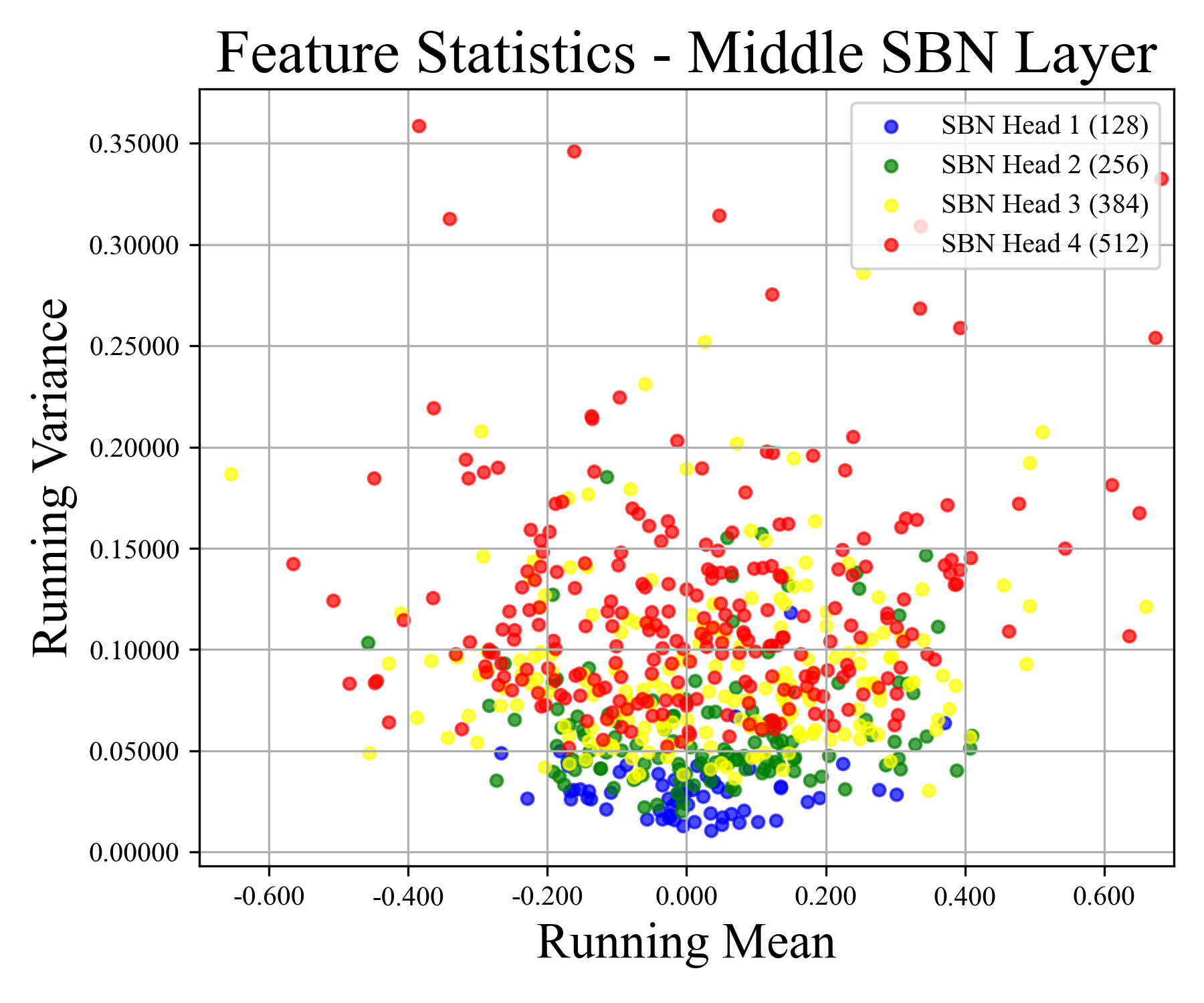}  \includegraphics[width=0.7\linewidth]{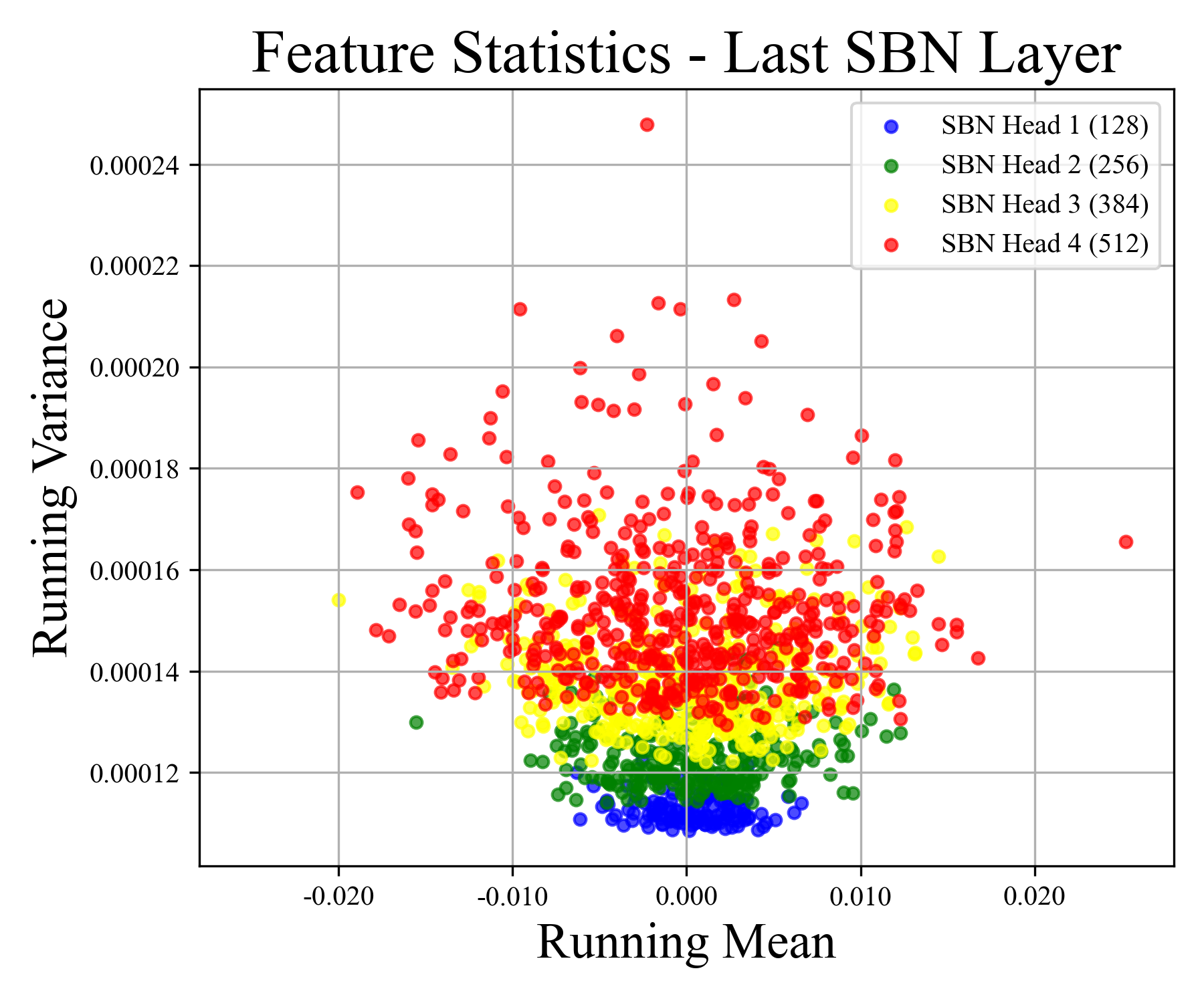}
  \caption{Visualization of running means and variances of the Switchable Batch Norm (SBN) layers for clean data in OATS-based ResNet-34 (with four sub-nets) during training. Top plot shows scattered statistics in the middle SBN, while the bottom plot reveals tighter and more distinct distributions in the final SBN. The shared output layer in OATS causes representational interference. SOL alleviates this by assigning separate classification heads to the sub-nets, thereby improving calibration and performance without increasing their parameter count.}
  \vspace{-10pt}
  \label{fig:fig_3}
\end{figure}

\begin{equation}
    \min_{\Theta} \frac{1}{N} \sum_{i=1}^{N} \mathcal{L}(\mathcal{N}(x_i;\Theta),y_i)
\label{eq:eq_7}
\end{equation}

During training, a width multiplier \( \alpha_k \in \mathcal{W} \) is sampled to activate the corresponding sub-net \( \mathcal{S}_{\alpha_k} \), which produces intermediate features \(f_{\alpha_k} = \mathcal{S}_{\alpha_k}(x) \). All sub-nets share a common classification head \( \text{C} \), used to compute logits: \(z_{\alpha_k} = \text{C}(f_{\alpha_k}) \).

Since sub-nets of different widths produce different feature distributions \cite{yu2018slimmable, wang_once-for-all_2020}, SNNs \cite{yu2018slimmable} used SBN layers and OATS \cite{wang_once-for-all_2020} used SDBN layers to maintain separate normalization parameters for each sub-nets (and data type). Although these layers help decouple intermediate feature statistics, all sub-nets still converge at the shared output layer, which is commonly adopted in traditional OFA methods \cite{cai_once-for-all:_2020,chen2021onlytrainonce,yu2018slimmable,yu2019universally,kundu2023sparse,wang_once-for-all_2020}.
This shared head creates a bottleneck by forcing incompatible features into a single representation space, leading to degraded performance. The design space becomes more complex \cite{yu2018slimmable,yu2019universally, wang_once-for-all_2020, kundu_float:_2023}, when the number of sub-nets increases. To investigate this, we analyzed the statistics of the middle and final \( \mathrm{BN}_{\text{cln}} \) layers of ResNet-34 super-net in Figure~\ref{fig:fig_3}. The figure reveals significant variance across sub-nets: the narrowest sub-net (blue) exhibits compact, low-variance features, whereas the widest (red) shows high-variance, dispersed distribution. This indicates that, although SBN and SDBN preserve internal feature separation, the \textbf{shared output head} remains a bottleneck at the end of the pipeline.

\subsection{Overcoming Bottleneck with Switchable Output Layer (Separate Parameters)}

\begin{figure*}[!t]
    \centering    \includegraphics[width=0.7\linewidth]{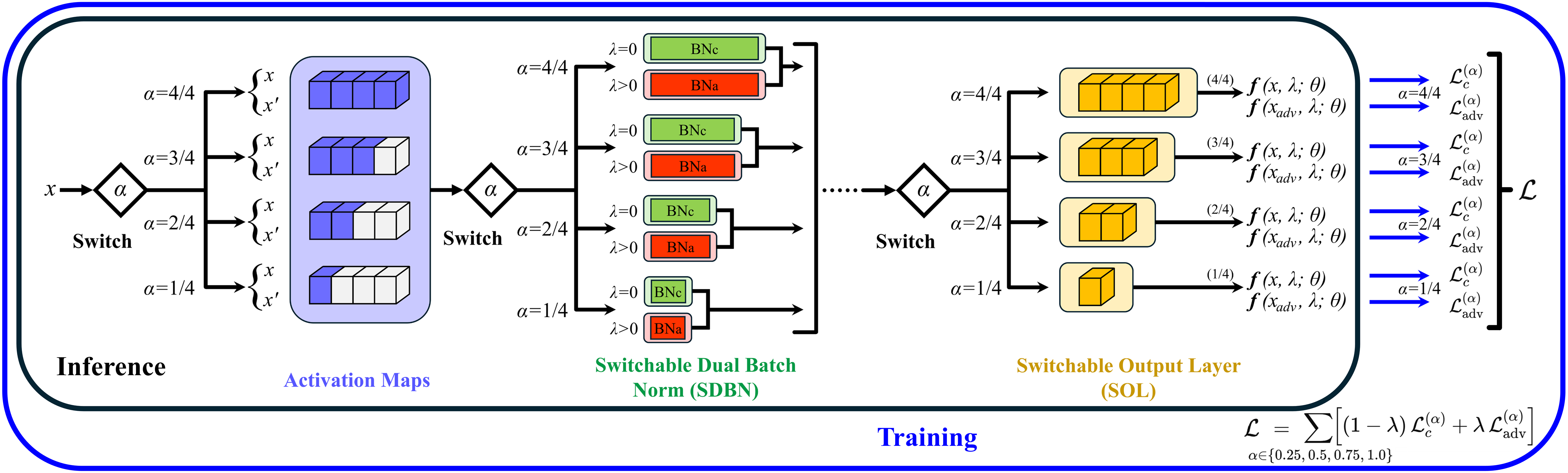}
    \caption{\centering Illustration of OATS framework \cite{wang_once-for-all_2020} with Switchable Dual Batch Norm (SDBN) layers and Switchable Output Layer (SOL). Value of $\alpha$ indicates the width fraction for the corresponding sub-net in the backbone that is activated during training or inference. Convolutional layers have shared parameters whereas SDBN and SOL have separate parameters for each sub-net.}
    \label{fig:fig_4}
\end{figure*}

As established in previous section, the shared output layer in OFA frameworks introduces a critical bottleneck by forcing diverse sub-net features into a single classification space, leading to interference and degraded performance. The \textbf{Switchable Output Layer (SOL)} enhances OFA frameworks \cite{yu2018slimmable, yu2019universally, cai_once-for-all:_2020, chen2021onlytrainonce, wang_once-for-all_2020, kundu2023sparse} by addressing performance degradation caused by the shared output layer across sub-nets. SOL introduces a lightweight modification by assigning a separate classification head to each sub-net in the super-net backbone, thereby preserving the sub-net specific representations through to the output and eliminating cross sub-net conflict during training (see Figure~\ref{fig:fig_2}).

Formally, instead of sharing a common classification head \( \text{C} \) for all sub-nets, SOL introduces a unique head \( \text{C}_{\alpha_k} \) for each sub-net \( \mathcal{S}_{\alpha_k} \). During training, a width multiplier \( \alpha_k \in \mathcal{W} \) is used to activate the corresponding sub-net to produce intermediate features:

\begin{equation}
z_{\alpha_k} = \text{C}_{\alpha_k}(f_{\alpha_k})
\label{eq:eq_8}
\end{equation}

Only the active head \( \text{C}_{\alpha_k} \) receives updates during backpropagation, while others remain inactive. This dynamic switching ensures that each sub-net learns independently, avoiding interference from incompatible gradients at the output layer (see Figure~\ref{fig:fig_2} and Figure~\ref{fig:fig_4}).

The full OATS-SOL pipeline is detailed in Algorithm~1. SOL is agnostic to the choice of loss function and supports a range of training objectives, including standard cross-entropy \(\mathcal{L}_{CE}\) \cite{lecun2015deep, krizhevsky2012imagenet} or distillation \(\mathcal{L}_{KLD}\) \cite{hinton2015distilling,sun2024logit,gou2021knowledge}, and robust objectives like TRADES \cite{zhang_theoretically_2019}, MMA \cite{ding2018mma}, and adversarial distillation \cite{goldblum2020adversarially,papernot_distillation_2016,gou2021knowledge}. For hybrid training (as in OATS), the total loss across eight sub-nets is defined as:
\begin{equation}
\mathcal{L} = \sum_{\alpha \in \{1/8,\, 2/8,\, ...,\, 8/8\}} \left[ (1 - \lambda) \, \mathcal{L}_{\text{CE}}^{(\alpha)} + \lambda \, \mathcal{L}_{\text{ADV}}^{(\alpha)} \right]
\label{eq:eq_9}
\end{equation}

\begin{algorithm}[t]
\caption{Once-for-All Adversarial Training and Slimming with Switchable Output Layer (OATS-SOL)}
\label{alg:outline}
\begin{algorithmic}[1]  
\Require Training set $\mathcal{D}$, set $\mathbb{S}$ with $\lambda$ values, Super-Net $\mathcal{N}$, max iterations $T$, width multipliers list $\mathcal{W}$
\Ensure Network parameters $\Theta$
\For{$t = 1$ to $T$}
    \State Sample batch $(x, y)$ from $\mathcal{D}$
    \State Sample $\lambda$ from 
    $\mathbb{S}$
    \State Clear gradients: $optimizer.zero\_grad()$
    \For{each \(\alpha\) in $\mathcal{W}$}
        \State Activate sub-net
        $\mathcal{S}_{\alpha_k}$ with head $\text{C}_{\alpha_k}$ in $\mathcal{N}$
        \State Generate adversarial example $x_{\text{adv}}$
        \State Compute loss $loss = \mathcal{L}(x, y, \lambda)$
        \State Accumulate gradients: $loss.backward()$
    \EndFor
    \State Update parameters: $optimizer.step()$
\EndFor
\end{algorithmic}
\end{algorithm}
\section{Experiments \& Results}
\label{sec:experiments}
\subsection{Super-Net Architectures and Setup}

We evaluate SOLAR on four super-net backbones: WideResNet-16-8 \cite{zagoruyko2016wideRN}, ResNet-34 \cite{he2016deepRN}, WideResNet-40-2 \cite{zagoruyko2016wideRN}, and MobileNetV2 \cite{sandler2018mobilenetv2}, using five benchmark datasets: SVHN \cite{netzer2011reading}, CIFAR-10 \cite{krizhevsky2009learning}, STL-10 \cite{coates2011analysis}, CIFAR-100 \cite{krizhevsky2009learning}, and TinyImageNet \cite{le2015tiny}. Experiments have been conducted on a workstation with Intel Core i9-14900KF CPU, 36 MB L3 cache, and NVIDIA RTX 4090. The code of \textbf{SOLAR} is publicly available at: \href{https://github.com/sakt90/SOLAR}{https://github.com/sakt90/SOLAR}.

\begin{figure*}[!t]
    \centering    \includegraphics[width=0.8\linewidth]{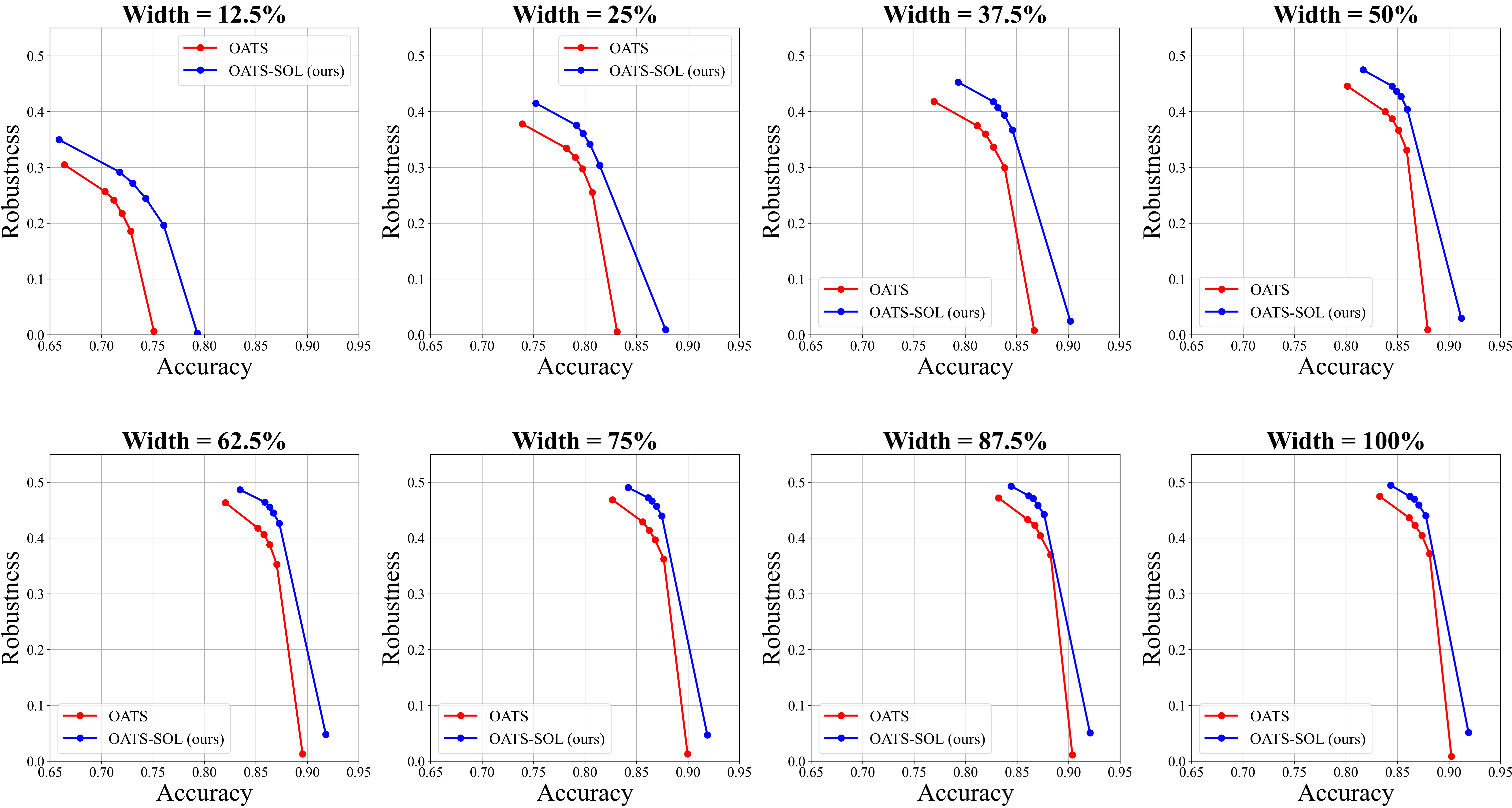}
    \caption{\centering Comparison of OATS \cite{wang_once-for-all_2020} and OATS-SOL on CIFAR-10 dataset using ResNet-34 backbone (with 8 sub-nets). OATS-SOL provides superior performance than OATS for all the sub-nets in terms of accuracy and PGD-7 based robustness.}
    \vspace{-5pt}
    \label{fig:fig_5}
\end{figure*}

\subsection{Hyperparameter Settings}
For OATS-SOL, we trained on SVHN, CIFAR-10, STL-10, and CIFAR-100 for 40, 120, 120, and 120 epochs, respectively. For SNN-SOL, training was conducted on CIFAR-10, CIFAR-100, and TinyImageNet for 120, 120, and 140 epochs. Batch sizes were set to 64 for STL-10 and TinyImageNet, and 128 for the other datasets. All experiments used SGD with momentum 0.9, a cosine annealing scheduler, and learning rates: \{0.1, 0.05, 0.01\}. Reported results correspond to the best performance across multiple runs with different random seeds, selected based on optimal validation accuracy.

\textbf{Adversarial Training and Evaluation:\quad} We use same settings as OATS \cite{wang_once-for-all_2020}. AT is performed using 7-step PGD attack ($L_{\infty}$ norm), $\epsilon=8/255$, and step-size $=2/255$. We evaluate the models on the basis of \textbf{Accuracy} and \textbf{Robustness}. Following OATS \cite{wang_once-for-all_2020}, we uniformly sample $\lambda$ element-wise from the set $\mathbb{S} = \{0.0, 0.1, 0.2, 0.3, 0.4, 1.0\}$ during training and validation. For inference, $\lambda$ can be any value in [0,1], as per requirements for the accuracy-robustness trade-off.

\begin{figure*}[!t]
    \centering    \includegraphics[width=0.8\linewidth]{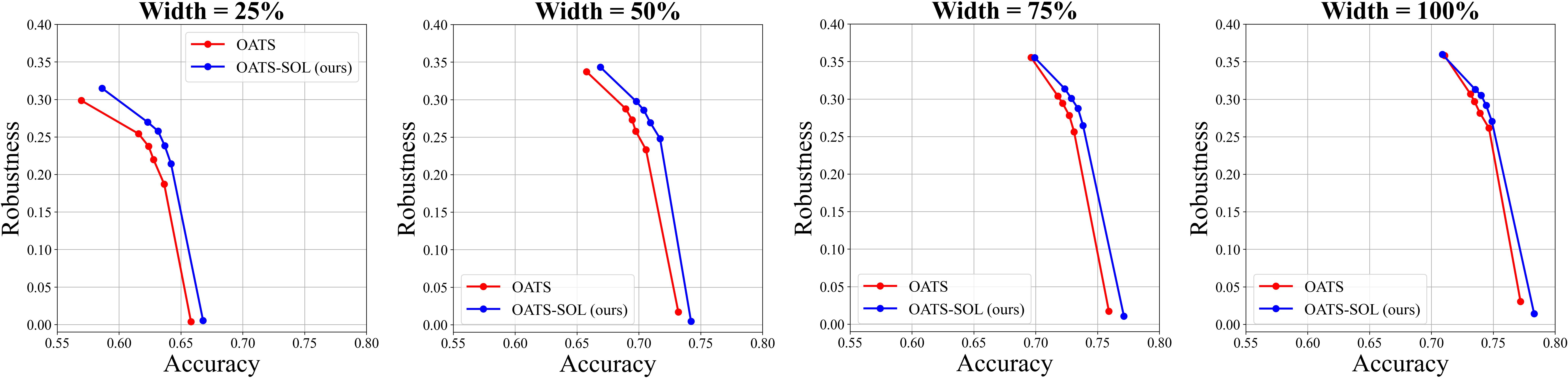}
    \caption{\centering Comparison of OATS \cite{wang_once-for-all_2020} and OATS-SOL on STL-10 dataset using WideResNet-40-2 as backbone with 4 sub-nets.}
    \vspace{-5pt}
    \label{fig:fig_6}
\end{figure*}

\begin{figure*}[!t]
    \centering    \includegraphics[width=0.8\linewidth]{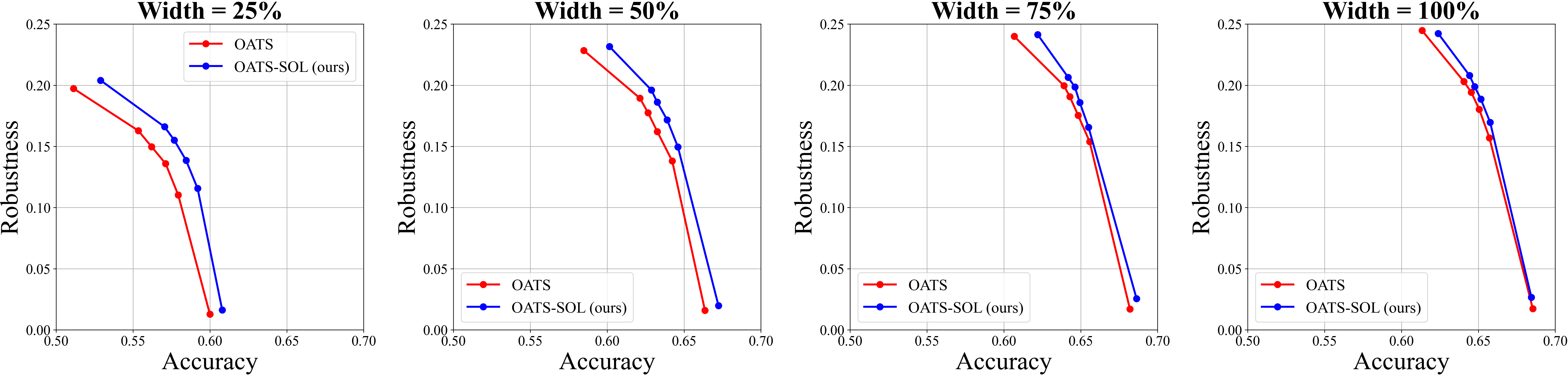}
    \caption{\centering Comparison of OATS \cite{wang_once-for-all_2020} and OATS-SOL on CIFAR-100 dataset using ResNet-34 as backbone with 4 sub-nets.}
    \label{fig:fig_7}
\end{figure*}

\begin{table*}[!t]
\definecolor{my_blue}{RGB}{0,0,255}
\centering
\caption{\centering Accuracy of SNN \cite{yu2018slimmable} and SNN-SOL based Sub-Nets on CIFAR-10 for backbones packed with 16 Sub-Nets.}
\vspace{-3pt}
\scriptsize
\begin{tabular}{|>{\centering\arraybackslash}p{0.9cm}|>{\centering\arraybackslash}p{0.7cm}|>{\centering\arraybackslash}p{1.1cm}|>{\centering\arraybackslash}p{1.0cm}|>{\centering\arraybackslash}p{0.7cm}|>{\centering\arraybackslash}p{1.1cm}|>{\centering\arraybackslash}p{1.0cm}|>{\centering\arraybackslash}p{0.7cm}|>{\centering\arraybackslash}p{1.1cm}|>
{\centering\arraybackslash}p{1.0cm}|}
\hline
\multirow{2}{*}{\scriptsize \shortstack{\textbf{\rule{0pt}{0.75em}Sub-Net}\\\textbf{\rule{0pt}{0.75em}Width}}} & \multicolumn{3}{|c|}{\footnotesize \textbf{\rule{0pt}{0.9em}ResNet-34}} & \multicolumn{3}{|c|}{\footnotesize \textbf{\rule{0pt}{0.9em}WideResNet-16-8}} & \multicolumn{3}{|c|}{\footnotesize \textbf{\rule{0pt}{0.9em}MobileNetV2}} \\
\cline{2-10}
 & {\scriptsize \rule{0pt}{1.0em} SNN} & {\scriptsize \rule{0pt}{1.0em} SNN-SOL} & {\scriptsize \rule{0pt}{1.0em} Gain \% $\uparrow$ } & {\scriptsize \rule{0pt}{1.0em} SNN} & {\scriptsize \rule{0pt}{1.0em} SNN-SOL} & {\scriptsize \rule{0pt}{1.0em} Gain \% $\uparrow$ } & {\scriptsize \rule{0pt}{1.0em} SNN} & {\scriptsize \rule{0pt}{1.0em} SNN-SOL} & {\scriptsize \rule{0pt}{1.0em} Gain \% $\uparrow$ } \\
\hline
\multicolumn{10}{|c|}{\footnotesize \textbf{\rule{0pt}{0.9em} Small Sub-Nets}} \\
\hline
1/16 & 83.28 & 83.55 & \textbf{0.27} & 82.72 & 82.94 & \textbf{0.22} & 82.16 & 84.17 & \textbf{2.01} \\
2/16 & 87.71 & 88.01 & \textbf{0.30} & 86.96 & 88.37 & \textbf{1.41} & 86.24 & 88.91 & \textcolor{my_blue}{\textbf{2.67}} \\
3/16 & 89.88 & 90.07 & \textbf{0.19} & 88.98 & 91.41 & \textcolor{my_blue}{\textbf{2.43}} & 88.13 & 90.50 & \textbf{2.37} \\
4/16 & 90.96 & 91.27 & \textbf{0.31} & 92.35 & 92.46 & \textbf{0.11} & 89.18 & 90.59 & \textbf{1.41} \\
5/16 & 91.86 & 91.88 & \textbf{0.02} & 92.51 & 93.12 & \textbf{0.61} & 89.56 & 91.96 & \textbf{2.40} \\
6/16 & 92.32 & 92.64 & \textcolor{my_blue}{\textbf{0.32}} & 93.00 & 93.63 & \textbf{0.63} & 90.26 & 92.61 & \textbf{2.35} \\
\hline
\multicolumn{10}{|c|}{\footnotesize \textbf{\rule{0pt}{0.9em} Medium Sub-Nets}} \\
\hline
7/16 & 92.48 & 92.98 & \textbf{0.50} & 93.14 & 93.81 & \textcolor{blue}{\textbf{0.67}} & 90.82 & 92.70 & \textbf{1.88} \\
8/16 & 92.54 & 93.35 & \textbf{0.81} & 93.78 & 93.95 & \textbf{0.17} & 90.68 & 92.88 & \textcolor{blue}{\textbf{2.20}} \\
9/16 & 92.78 & 93.48 & \textbf{0.70} & 93.93 & 94.23 & \textbf{0.30} & 91.37 & 92.87 & \textbf{1.50} \\
10/16 & 92.84 & 93.80 & \textbf{0.96} & 93.97 & 94.20 & \textbf{0.23} & 91.80 & 93.18 & \textbf{1.38} \\
11/16 & 92.88 & 93.86 & \textcolor{blue}{\textbf{0.98}} & 93.91 & 94.28 & \textbf{0.37} & 91.51 & 93.38 & \textbf{1.87} \\
\hline
\multicolumn{10}{|c|}{\footnotesize \textbf{\rule{0pt}{0.9em} Large Sub-Nets}} \\
\hline
12/16 & 93.07 & 93.84 & \textcolor{blue}{\textbf{0.77}} & 94.12 & 94.46 & \textbf{0.34} & 91.91 & 93.21 & \textbf{1.30} \\
13/16 & 93.13 & 93.89 & \textbf{0.76} & 94.07 & 94.55 & \textbf{0.48} & 91.93 & 93.29 & \textbf{1.36} \\
14/16 & 93.27 & 93.98 & \textbf{0.71} & 94.11 & 94.62 & \textbf{0.51} & 92.12 & 93.64 & \textbf{1.52} \\
15/16 & 93.32 & 93.94 & \textbf{0.62} & 94.10 & 94.57 & \textbf{0.47} & 92.03 & 93.65 & \textcolor{blue}{\textbf{1.62}} \\
16/16 & 93.38 & 94.04 & \textbf{0.66} & 94.09 & 94.67 & \textcolor{blue}{\textbf{0.58}} & 92.15 & 93.45 & \textbf{1.30} \\
\hline
\end{tabular}
\label{tab:table_1}
\end{table*}

\begin{table*}[!t]
\definecolor{my_blue}{RGB}{0,0,255}
\centering
\caption{\centering Accuracy of SNN \cite{yu2018slimmable} and SNN-SOL on CIFAR-100 for backbones with 8 sub-nets.}
\vspace{-3pt}
\scriptsize
\begin{tabular}{|>{\centering\arraybackslash}p{0.9cm}|>{\centering\arraybackslash}p{0.7cm}|>{\centering\arraybackslash}p{1.1cm}|>{\centering\arraybackslash}p{1.0cm}|>{\centering\arraybackslash}p{0.7cm}|>{\centering\arraybackslash}p{1.1cm}|>{\centering\arraybackslash}p{1.0cm}|>{\centering\arraybackslash}p{0.7cm}|>{\centering\arraybackslash}p{1.1cm}|>
{\centering\arraybackslash}p{1.0cm}|}
\hline
\multirow{2}{*}{\scriptsize \shortstack{\textbf{\rule{0pt}{0.75em}Sub-Net}\\\textbf{\rule{0pt}{0.75em}Width}}} & \multicolumn{3}{|c|}{\footnotesize \textbf{\rule{0pt}{0.9em}ResNet-34}} & \multicolumn{3}{|c|}{\footnotesize \textbf{\rule{0pt}{0.9em}WideResNet-16-8}} & \multicolumn{3}{|c|}{\footnotesize \textbf{\rule{0pt}{0.9em}MobileNetV2}} \\
\cline{2-10}
 & {\scriptsize \rule{0pt}{1.0em} SNN} & {\scriptsize \rule{0pt}{1.0em} SNN-SOL} & {\scriptsize \rule{0pt}{1.0em} Gain \% $\uparrow$ } & {\scriptsize \rule{0pt}{1.0em} SNN} & {\scriptsize \rule{0pt}{1.0em} SNN-SOL} & {\scriptsize \rule{0pt}{1.0em} Gain \% $\uparrow$ } & {\scriptsize \rule{0pt}{1.0em} SNN} & {\scriptsize \rule{0pt}{1.0em} SNN-SOL} & {\scriptsize \rule{0pt}{1.0em} Gain \% $\uparrow$ } \\
\hline
1/8 & 66.23 & 66.18 & \textbf{-0.05} & 64.54 & 64.33 & \textbf{-0.21} & 64.45 & 64.37 & \textbf{-0.08} \\
2/8 & 71.64 & 72.07 & \textbf{0.43} & 71.64 & 71.96 & \textbf{0.32} & 68.94 & 68.91 & \textbf{-0.03} \\
3/8 & 73.91 & 74.82 & \textbf{0.91} & 74.48 & 74.28 & \textbf{-0.20} & 71.16 & 71.59 & \textbf{0.43} \\
4/8 & 75.33 & 76.26 & \textbf{0.93} & 75.86 & 75.97 & \textbf{0.11} & 72.15 & 72.38 & \textbf{0.23} \\
5/8 & 76.26 & 77.28 & \textbf{1.02} & 76.71 & 76.93 & \textbf{0.22} & 73.04 & 73.54 & \textbf{0.50} \\
6/8 & 76.79 & 78.10 & \textbf{1.31} & 76.91 & 77.27 & \textbf{0.36} & 73.26 & 73.78 & \textbf{0.52} \\
7/8 & 76.74 & 78.08 & \textbf{1.34} & 77.37 & 77.83 & \textcolor{my_blue}{\textbf{0.46}} & 73.51 & 73.92 & \textbf{0.41} \\
8/8 & 76.78 & 78.43 & \textcolor{my_blue}{\textbf{1.65}} & 77.97 & 78.22 & \textbf{0.25} & 73.23 & 74.19 & \textcolor{my_blue}{\textbf{0.96}} \\
\hline
\end{tabular}
\label{tab:table_2}
\end{table*}

\begin{table*}[!t]
\definecolor{my_blue}{RGB}{0,0,255}
\centering
\caption{\centering Accuracy of SNN \cite{yu2018slimmable} and SNN-SOL on TinyImageNet for backbones with 8 sub-nets.}
\vspace{-3pt}
\scriptsize
\begin{tabular}{|>{\centering\arraybackslash}p{0.9cm}|>{\centering\arraybackslash}p{0.7cm}|>{\centering\arraybackslash}p{1.1cm}|>{\centering\arraybackslash}p{1.0cm}|>{\centering\arraybackslash}p{0.7cm}|>{\centering\arraybackslash}p{1.1cm}|>{\centering\arraybackslash}p{1.0cm}|>{\centering\arraybackslash}p{0.7cm}|>{\centering\arraybackslash}p{1.1cm}|>
{\centering\arraybackslash}p{1.0cm}|}
\hline
\multirow{2}{*}{\scriptsize \shortstack{\textbf{\rule{0pt}{0.75em}Sub-Net}\\\textbf{\rule{0pt}{0.75em}Width}}} & \multicolumn{3}{|c|}{\footnotesize \textbf{\rule{0pt}{0.9em}ResNet-34}} & \multicolumn{3}{|c|}{\footnotesize \textbf{\rule{0pt}{0.9em}WideResNet-16-8}} & \multicolumn{3}{|c|}{\footnotesize \textbf{\rule{0pt}{0.9em}MobileNetV2}} \\
\cline{2-10}
 & {\scriptsize \rule{0pt}{1.0em} SNN} & {\scriptsize \rule{0pt}{1.0em} SNN-SOL} & {\scriptsize \rule{0pt}{1.0em} Gain \% $\uparrow$ } & {\scriptsize \rule{0pt}{1.0em} SNN} & {\scriptsize \rule{0pt}{1.0em} SNN-SOL} & {\scriptsize \rule{0pt}{1.0em} Gain \% $\uparrow$ } & {\scriptsize \rule{0pt}{1.0em} SNN} & {\scriptsize \rule{0pt}{1.0em} SNN-SOL} & {\scriptsize \rule{0pt}{1.0em} Gain \% $\uparrow$ } \\
\hline
1/8 & 51.22 & 52.33 & \textbf{1.11} & 45.81 & 46.54 & \textbf{0.73} & 48.64 & 49.57 & \textbf{0.93} \\
2/8 & 57.79 & 59.65 & \textbf{1.86} & 56.30 & 56.22 & \textbf{-0.08} & 54.85 & 55.76 & \textbf{0.91} \\
3/8 & 60.12 & 61.73 & \textbf{1.61} & 60.02 & 60.26 & \textbf{0.24} & 57.62 & 58.97 & \textcolor{my_blue}{\textbf{1.35}} \\
4/8 & 61.62 & 63.41 & \textbf{1.79} & 61.93 & 62.86 & \textbf{0.93} & 58.67 & 59.37 & \textbf{0.70} \\
5/8 & 62.02 & 64.37 & \textbf{2.35} & 62.24 & 63.74 & \textbf{1.50} & 59.75 & 60.04 & \textbf{0.29} \\
6/8 & 62.83 & 65.65 & \textbf{2.82} & 63.34 & 64.73 & \textbf{1.39} & 59.67 & 60.31 & \textbf{0.64} \\
7/8 & 63.50 & 66.08 & \textbf{2.58} & 63.12 & 65.46 & \textcolor{my_blue}{\textbf{2.34}} & 59.66 & 60.81 & \textbf{1.15} \\
8/8 & 63.46 & 66.39 & \textcolor{my_blue}{\textbf{2.93}} & 63.61 & 65.34 & \textbf{1.73} & 59.91 & 61.17 & \textbf{1.26} \\
\hline
\end{tabular}
\vspace{-3pt}
\label{tab:table_3}
\end{table*}

\begin{table*}[!t]
\centering
\renewcommand{\arraystretch}{1.2}
\caption{\centering Parameter counts of SNN \cite{yu2018slimmable} vs. SNN-SOL backbones packed with 8, 32, and 64 sub-nets for CIFAR-10 dataset.}
\vspace{-3pt}
\scriptsize
\begin{tabular}{|>{\centering\arraybackslash}p{1.4cm}|>{\centering\arraybackslash}p{0.7cm}|>{\centering\arraybackslash}p{1.1cm}|>{\centering\arraybackslash}p{0.85cm}|>{\centering\arraybackslash}p{0.7cm}|>{\centering\arraybackslash}p{1.1cm}|>{\centering\arraybackslash}p{0.85cm}|>{\centering\arraybackslash}p{0.7cm}|>{\centering\arraybackslash}p{1.1cm}|>
{\centering\arraybackslash}p{0.85cm}|}
\hline
\multirow{2}{*}{\scriptsize \textbf{\rule{0pt}{1.0em}Super-Net}} & \multicolumn{3}{|c|}{\footnotesize \textbf{\rule{0pt}{0.9em}8 Sub-Nets}} & \multicolumn{3}{|c|}{\footnotesize \textbf{\rule{0pt}{0.9em}32 Sub-Nets}} & \multicolumn{3}{|c|}{\footnotesize \textbf{\rule{0pt}{0.9em}64 Sub-Nets}} \\
\cline{2-10}
 & {\scriptsize \rule{0pt}{1.0em} SNN} & {\scriptsize \rule{0pt}{1.0em} SNN-SOL} & {\scriptsize Increase} & {\scriptsize \rule{0pt}{1.0em} SNN} & {\scriptsize \rule{0pt}{1.0em} SNN-SOL} & {\scriptsize Increase} & {\scriptsize \rule{0pt}{1.0em} SNN} & {\scriptsize \rule{0pt}{1.0em} SNN-SOL} & {\scriptsize Increase} \\
\hline
\textbf{WRN-40-2} & 2.26M & 2.28M & 0.97\% & 2.33M & 2.41M & 3.61\% & 2.41M & 2.58M & 6.88\% \\
\textbf{MobileNetV2}    & 2.35M & 2.40M & 1.91\% & 2.75M & 2.96M & 7.23\% & 3.28M & 3.68M & 12.32\% \\
\textbf{WRN-16-8} & 10.99M & 11.00M & 0.16\% & 11.07M & 11.15M & 0.72\% & 11.19M & 11.35M & 1.45\% \\
\textbf{ResNet-34} & 21.34M & 21.36M & 0.08\% & 21.55M & 21.63M & 0.37\% & 21.82M & 21.98M & 0.74\% \\
\hline
\end{tabular}
\vspace{-3pt}
\label{tab:table_4}
\end{table*}

\subsection{Baseline Methods: OATS and SNNs} We evaluate SOLAR on two baseline: Once-for-All Adversarial Training and Slimming (OATS) \cite{wang_once-for-all_2020} and Slimmable Neural Networks (SNNs) \cite{yu2018slimmable}. After employing Switchable Output Layer (SOL), we denote them as OATS-SOL and SNN-SOL. For fair comparison, we evaluate the baseline and SOLAR frameworks using the same hyperparameters.

\textbf{Comparison with OATS: \quad} For comparing OATS and OATS-SOL, we used WideResNet-16-8, ResNet-34, WideResNet-40-2, and ResNet-34 as backbones on the SVHN, CIFAR-10, STL-10, and CIFAR-100 datasets, respectively. The accuracy and robustness of OATS and OATS-SOL sub-nets can be tuned by changing the values of $\lambda \in [0.0, 1.0]$ for free during run-time \cite{wang_once-for-all_2020}. To meet the tighter memory and storage constraints of devices, sub-nets with smaller widths are preferred. The comparison of OATS \cite{wang_once-for-all_2020} and OATS-SOL on SVHN dataset using WideResNet-16-8 backbone (packed with 8 sub-nets) is shown in Figure~\ref{fig:fig_1}. OATS-SOL is represented by \textbf{blue color} and provides superior performance than OATS for all the sub-nets in terms of accuracy and robustness. The performance gain is generally higher in smaller sub-nets as compared to larger ones. Similarly, Figure~\ref{fig:fig_5}, Figure~\ref{fig:fig_6}, and Figure~\ref{fig:fig_7} show the comparison of OATS and OATS-SOL on CIFAR-10 (ResNet-34 backbone with 8 sub-nets), STL-10 (WideResNet-40-2 backbone with 4 sub-nets), and CIFAR-100 (ResNet-34 backbone with 4 sub-nets), respectively. OATS-SOL provides better accuracy and robustness for all the sub-nets across the datasets. We present a numerical comparison between OATS and OATS-SOL for CIFAR-10 dataset in Table~\ref{tab:table_6} in the Appendix. In addition, the Appendix includes further insightful results supported by additional figures.

\textbf{Comparison with SNNs: \quad} We compare SNN~\cite{yu2018slimmable} and SNN-SOL using ResNet-34, WideResNet-16-8, and MobileNetV2 backbones on CIFAR-10, CIFAR-100, and TinyImageNet. Table~\ref{tab:table_1} reports results on CIFAR-10 with 16 sub-nets, where SNN-SOL achieves maximum gains of \textbf{0.32\%}, \textbf{2.43\%}, and \textbf{2.67\%} for small sub-nets, \textbf{0.98\%}, \textbf{0.67\%}, and \textbf{2.20\%} for medium sub-nets, and \textbf{0.77\%}, \textbf{0.58\%}, and \textbf{1.62\%} for large sub-nets across the three backbones. On CIFAR-100 (Table~\ref{tab:table_2}), SNN-SOL improves accuracies by up to \textbf{1.65\%}, \textbf{0.46\%}, and \textbf{0.96\%}, using the backbones with 8 sub-nets. On TinyImageNet (Table~\ref{tab:table_3}), improvements reach \textbf{2.93\%}, \textbf{2.34\%}, and \textbf{1.35\%} for ResNet-34, WideResNet-16-8, and MobileNetV2, all packed with 8 sub-nets, respectively. The experimental results demonstrate that when the super-net backbones are packed with higher number of sub-nets, the Switchable Output Layer (SOL) provides notable performance gains for the sub-nets.

\subsection{Impact of SOL on Parameter Count and FLOPs}
The training and inference overhead using SOL is negligible: only the active head is used for each sub-net, containing the same number of parameters as in the baseline. The shared output layer performs \textbf{\enquote{slimming}} whereas SOL performs \textbf{\enquote{switching}}. SOL increases parameter storage due to use of multiple heads in the super-net but does not increase the parameters for any sub-net. The performance gains come from the unshared nature of the parameters. Table~\ref{tab:table_4} shows the parameter increase due to SOL in the backbones. FLOPs during a forward pass for a SOL based sub-net with width fraction $\alpha_k$ is the sum of the feature encoder FLOPs $\mathcal{S}_{\alpha_k}$ and the FLOPs of its head $\text{C}_{\alpha_k}$, as in \eqref{eq:eq_10}.

\begin{equation}
\mathcal{F}\!\left(\mathcal{S}_{\alpha_k}\right) + \mathcal{F}\!\left(\mathcal{C}_{\alpha_k}\right) = \mathcal{F}\!\left(\mathcal{S}_{\alpha_k}\right) + \mathcal{F}\!\left(\mathcal{C}\right)
\label{eq:eq_10}
\end{equation}

Since, the number of parameters is same in the output layer after slimming or switching, FLOPs of the SOL sub-nets are identical to the corresponding baseline sub-nets, as shown in Table~\ref{tab:table_5}.

\begin{table}[h!]
\centering
\renewcommand{\arraystretch}{1.2}
\caption{\centering Identical FLOPs of SNN \cite{yu2018slimmable} and SNN-SOL sub-nets (with different widths), show that SOL adds no training overhead.}
\vspace{-3pt}
\scriptsize
\begin{tabular}{|c|c|c|c|c|c|}
\hline
\multirow{2}{*}{\textbf{Super-Net}} & \multicolumn{2}{c|}{\footnotesize \textbf{Sub-Net 1}} & \multicolumn{2}{c|}{\footnotesize \textbf{Sub-Net 4}} \\
\cline{2-5}
 & \scriptsize {SNN} & \scriptsize {SNN-SOL} & \scriptsize {SNN} & \scriptsize {SNN-SOL} \\
\hline
\textbf{WRN-40-2}    & 5,343,392 & 5,343,392 & 82,715,264 & 82,715,264 \\
\textbf{MobileNetV2} & 18,904,576 & 18,904,576 & 92,543,488 & 92,543,488 \\
\textbf{WRN-16-8}    & 24,472,192 & 24,472,192 & 388,082,176 & 388,082,176 \\
\textbf{ResNet-34}   & 18,565,760 & 18,565,760 & 291,318,272 & 291,318,272 \\
\hline
\end{tabular}
\label{tab:table_5}
\end{table}

\subsection{Post-Search Fine-Tuning of Sub-Nets}
We perform post-search fine-tuning on randomly selected sub-nets from the SNN and SNN-SOL backbones. As shown in Table~\ref{tab:table_6}, SOL sub-nets consistently achieve higher performance even after fine-tuning. This indicates that the unshared output heads in SOL enable sub-nets to learn stronger representations from the beginning. Fine-tuning refines these representations but does not fundamentally alter them, so well-optimized sub-nets continue to outperform. In contrast, sub-nets from the backbones with shared output layer cannot close this gap, as their representational limitations persist despite fine-tuning. Overall, SOL facilitates better optimization across all sub-nets by guiding them toward flatter, more generalizable minima, ensuring their dominance is preserved even after fine-tuning.

\begin{table}[h!]
\definecolor{my_blue}{RGB}{0,0,255}
\centering
\renewcommand{\arraystretch}{1.2}
\caption{\centering Performance of different sub-nets after fine-tuning.}
\vspace{-3pt}
\scriptsize
\begin{tabular}
{|>{\centering\arraybackslash}p{1.45cm}|>{\centering\arraybackslash}p{1.75cm}|>{\centering\arraybackslash}p{0.5cm}|>{\centering\arraybackslash}p{1.25cm}|>
{\centering\arraybackslash}p{1.15cm}|}
\hline
\textbf{\footnotesize Dataset} & \textbf{\footnotesize Sub-Net} & \textbf{\footnotesize SNN} & \textbf{\footnotesize SNN-SOL} & \textbf{\footnotesize Gain \%} $\uparrow$ \\
\hline
\multirow{2}{*}{\textbf{CIFAR-10}} & MobileNetV2$_{2/16}$ & 87.17 & 89.59 & \textcolor{my_blue}{\textbf{2.42}} \\
                                      & WRN-16-8$_{2/16}$ & 87.62 & 89.20 & \textcolor{my_blue}{\textbf{1.58}} \\
\hline
\multirow{2}{*}{\textbf{CIFAR-100}} & MobileNetV2$_{8/8}$ & 73.41 & 74.38 & \textcolor{my_blue}{\textbf{0.97}} \\
                                      & ResNet-34$_{8/8}$ & 77.29 & 78.86 & \textcolor{my_blue}{\textbf{1.57}} \\
\hline
\multirow{2}{*}{\textbf{TinyImageNet}} & WRN-16-8$_{5/8}$ & 62.66 & 63.84 & \textcolor{my_blue}{\textbf{1.18}} \\
                                      & ResNet-34$_{5/8}$ & 62.34 & 64.53 & \textcolor{my_blue}{\textbf{2.19}} \\
\hline
\end{tabular}
\vspace{-15pt}
\label{tab:table_6}
\end{table}

\section{Conclusion and Future Directions}
We introduce \textbf{Switchable Output Layer (SOL)} to enhance the performance and robustness of Once-for-All (OFA) training frameworks. SOL assigns independent classification heads to the sub-nets in the super-net backbone, which decouples their logit learning processes, mitigating the competition at the \textit{shared output layer}—a bottleneck limiting the sub-net accuracy, robustness, and optimization. Extensive experiments on two different baseline methods: Once-for-All Adversarial Training and Slimming (OATS) and Slimmable Neural Networks (SNNs), across multiple datasets and diverse super-net architectures, demonstrate that incorporation of SOL consistently improves performance of sub-nets without introducing additional training overhead or complexity.
SOL generalizes well, which highlights its potential as an effective enhancement for the OFA frameworks, encouraging flexible, scalable, and reliable deployment of specialized models across a wider range of devices and constraints. For future work, we aim to extend SOL to more OFA frameworks (e.g. \cite{cao2023three}, \cite{yu2019universally}, \cite{zhao2025slimmable}) and conduct large-scale evaluations on the ImageNet-1K dataset. In addition, we intend to study the impact of layer normalization on reducing representational interference across the sub-nets.

{
    \small    \bibliographystyle{ieeenat_fullname}
    \bibliography{wacv}
}
\end{document}